\documentclass{article} 
\usepackage{iclr2026_conference,times}
\usepackage{graphicx}
\usepackage{subcaption}
\usepackage{comment}

\usepackage{amsmath,amsfonts,bm}

\def\eqref#1{equation~\ref{#1}}

\def\1{\bm{1}}

\DeclareMathAlphabet{\mathsfit}{\encodingdefault}{\sfdefault}{m}{sl}
\SetMathAlphabet{\mathsfit}{bold}{\encodingdefault}{\sfdefault}{bx}{n}

\usepackage[inline,shortlabels]{enumitem}

\usepackage{hyperref}
\usepackage{graphicx}
\usepackage{booktabs}
\usepackage{multirow}
\usepackage{makecell}
\usepackage{url}
\usepackage{float}

\title{Investigating simple target-covariate relationships for \texttt{Chronos-2} and \texttt{TabPFN-TS} }

\author{Gaspard Berthelier\thanks{Equal contribution.} \,\thanks{Inria Sophia-Antipolis, Université Côte d'Azur.} , Mariia Baranova\footnotemark[1] \,\thanks{Université Paris Cité.} , Andrei-Tiberiu Pantea,\\
\textbf{Etienne Le Naour, Adrien Petralia, Tahar Nabil} \\
EDF R\&D, Palaiseau\\
\texttt{\{gaspard.berthelier,mariia.baranova,andrei.pantea\}@edf.fr} \\
\texttt{\{etienne.le-naour,tahar.nabil,adrien.petralia\}@edf.fr} \\
\vspace{-7mm} \AND 
Themis Palpanas \\
{Université Paris Cité} \\
\texttt{themis@mi.parisdescartes.fr} \\
}

\iclrfinalcopy

\track{Industry \& Application}

\begin{document}

\maketitle

\vspace{-4mm}
\begin{abstract}

Time Series Foundation Models (TSFMs) have recently achieved state-of-the-art performance, often outperforming supervised models in zero-shot settings. 
Recent TSFM architectures, such as \texttt{Chronos-2} and \texttt{TabPFN-TS}, aim to integrate covariates. In this paper, we design controlled experiments based on simple target–covariate relationships to assess this integration capability. Our results show that \texttt{TabPFN-TS} captures these relationships more effectively than \texttt{Chronos-2}, especially for short horizons, suggesting that the strong benchmark performance of \texttt{Chronos-2} does not automatically translate into optimal modeling of simple covariate–target dependencies.
\end{abstract}

\vspace{-4mm}
\section{Introduction}
\vspace{-2mm}

Time Series Foundation Models (TSFMs) enable inference-only forecasting leveraging massive pretraining on large-scale real-world and synthetic datasets. Extensive evaluations in \texttt{GiftEval} \citep{aksu2024gift} show that TSFMs outperform supervised deep learning baselines on univariate forecasting tasks from various domains. 

However, in real-world applications, univariate forecasting is often insufficient, and effectively leveraging dynamic covariates becomes essential, particularly in domains such as energy or retail.
The first generation of TSFMs \citep{TimesFM, Chronosv1} lacked native covariate handling, with the exception of \texttt{MOIRAI} \citep{woo2024unified} which was superseded by \texttt{MOIRAI 2.0} for improved univariate performance \citep{liu2025moirai}. 
More recently, \texttt{Chronos-2} \citep{ansari2024chronos2} and \texttt{TabPFN-TS} \citep{hoo2024tables2time} have shown impressive results on covariate-based forecasting, as reflected by \texttt{fev-bench} \citep{shchur2025fev}. 
To handle covariate integration, \texttt{Chronos-2} uses time and group attention layers to facilitate information exchange across multiple time series and is pretrained on synthetic multivariate data.
In contrast, \texttt{TabPFN} \citep{hollmann2022tabpfn} is a tabular foundation model pretrained on millions of synthetic regression tasks, which can be adapted for time series forecasting with carefully designed features \citep{hoo2024tables2time}. Both models leverage in-context learning to capture feature–target relationships on unseen problems.
Despite \texttt{Chronos-2}'s strong performance in recent bencharks (e.g. \texttt{fev-bench}), these evaluations offer limited insight into actual covariate usage.

In this paper, we investigate how effectively these two models leverage covariate information, focusing on cases where the relationship between the target and covariates follows simple patterns. Our experimental results show that \texttt{TabPFN-TS} more reliably captures simple target–covariate relationships
particularly for short forecasting horizons. These results suggest that, despite \texttt{Chronos-2}’s dominance on large-scale covariate-based forecasting benchmarks,
there is still significant potential to design TSFMs that more effectively leverage covariates.

\section{Experimental Evaluation}

In the following experiments, we compare \texttt{Chronos-2} and \texttt{TabPFN-TS} on minimal tasks designed to isolate target and covariate dependency. We compare the two models on a range of real-world datasets and forecasting settings, as well as on synthetically generated time series.
\paragraph{Settings.} We consider the usual time series forecasting setting, where models take as input a target's past look-back window of size $L$, to forecast a future horizon window of size $H$. Additionally, covariates can be included as inputs, throughout the full context (look-back and horizon). We build four different experiments: \begin{enumerate*}[(i)] \item Identity: the covariate equals the target to forecast; performed on real-world time series.
\item Sum: the target is $Z = X + Y$, with $X$ and $Y$ synthetically generated time series.
\item Aggregate: the target is an aggregate of the form $Z=\sqrt{X}+Y-Z^2$, where $X,Y,Z$ are real-world time series.
\item Quadratic: the target is of the form $Z = a + bX + cY^2$, with $X$ and $Y$ synthetically generated time series. More details on the experiment protocols are provided in Appendix~\ref{apx:protocol}.
\end{enumerate*}
 
\paragraph{Results for the identity experiment.} 
Table~\ref{tab:identity}, reports the errors of each model when the target is also given as a covariate. 
We observe that (i) while \texttt{Chronos-2} is a stronger univariate forecaster, \texttt{TabPFN-TS} consistently outperforms it in this experiment; (ii) despite being given access to the ground truth, both models still suffer from very short context lengths.

\begin{table}[h!]
\caption{Normalized MSEs for \texttt{Chronos-2} and \texttt{TabPFN-TS} before and after providing the target as covariate (+id), on real-world datasets and a range of $L-H$ settings.}
\vspace{1mm}
\centering
\scalebox{0.6}{
\begin{tabular}{l|ccc|ccc}
\toprule
 Dataset & Setting
 & \thead{\texttt{Chronos-2}}
 & \thead{\texttt{Chronos-2}(+id)}
 & \thead{\texttt{TabPFN-TS}}
 & \thead{\texttt{TabPFN-TS}(+id)} \\
\midrule
\multirow{4}{*}{\textsc{Electricity}} & 1344-336 & 0.2037 & 0.0085 & 0.2367 & \textbf{0.0056} \\
 & 672-168 & 0.1788 & \textbf{0.0130} & 0.2149 & 0.0205 \\
 & 168-24 & 0.1953 & 0.0477 & 0.2662 & \textbf{0.0415} \\
 & 24-24 & 1.3718 & 1.0252 & 2.1728 & \textbf{0.8121} \\
\midrule
\multirow{4}{*}{\textsc{Solar}} & 1344-336 & 0.1503 & 0.0456 & 0.2421 & \textbf{0.0000} \\
 & 672-168 & 0.1831 & 0.0612 & 0.1925 & \textbf{0.0000} \\
 & 168-24 & 0.1272 & 0.0886 & 0.1434 & \textbf{0.0000} \\
 & 24-24 & 1.3238 & 0.8333 & 1.8876 & \textbf{0.3565} \\
\midrule
\multirow{4}{*}{\textsc{Traffic}} & 1344-336 & 0.2326 & 0.0276 & 0.2893 & \textbf{0.0037} \\
 & 672-168 & 0.2158 & 0.0251 & 0.2550 & \textbf{0.0051} \\
 & 168-24 & 0.1926 & 0.0561 & 0.4007 & \textbf{0.0034} \\
 & 24-24 & 1.7487 & 1.5570 & 3.1238 & \textbf{1.1547} \\
\bottomrule
\end{tabular}
}
\label{tab:identity}
\end{table}

\begin{figure}[H]
    \centering
    \includegraphics[width=0.45\textwidth]{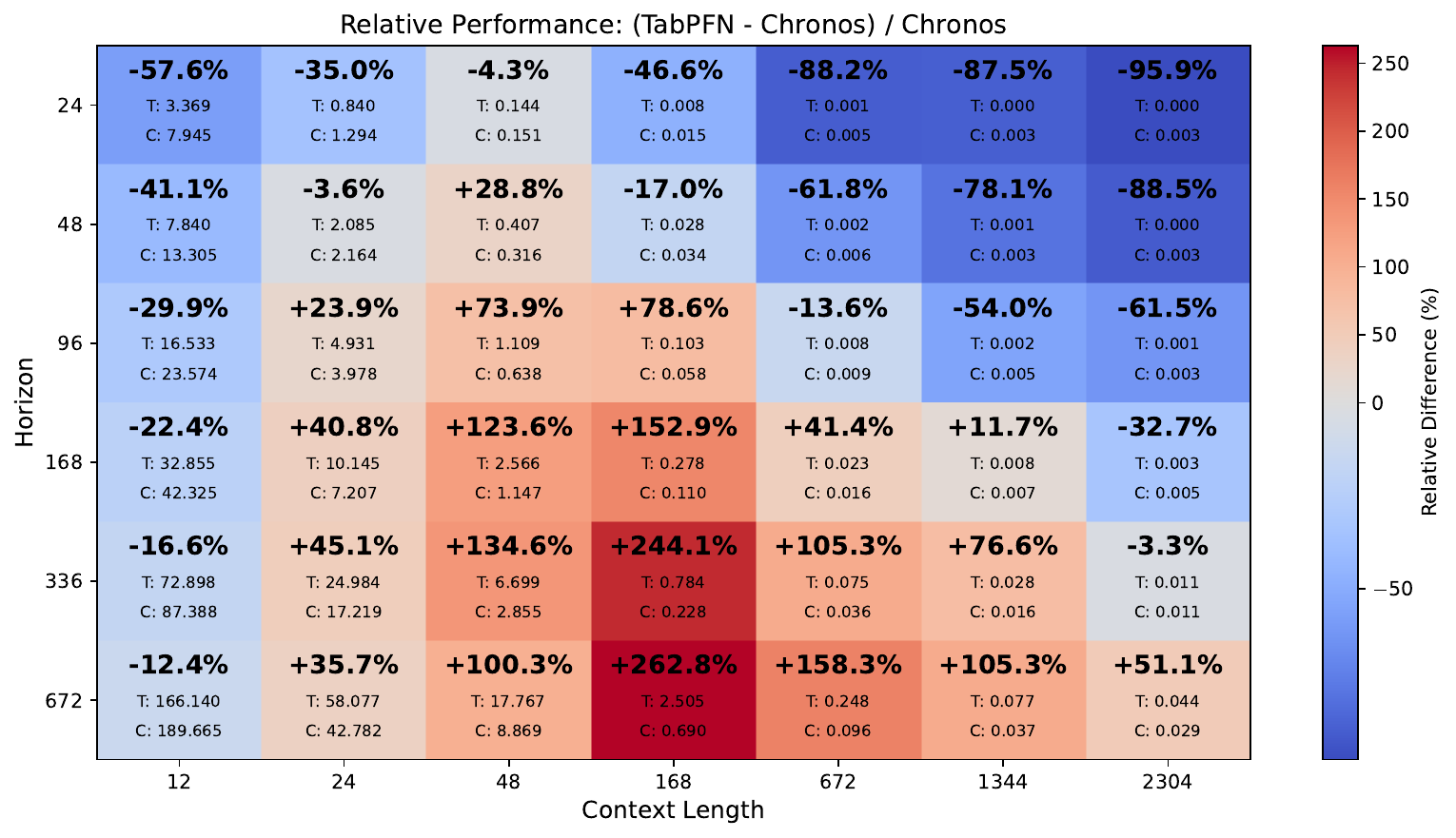}
    \includegraphics[width=0.45\textwidth]{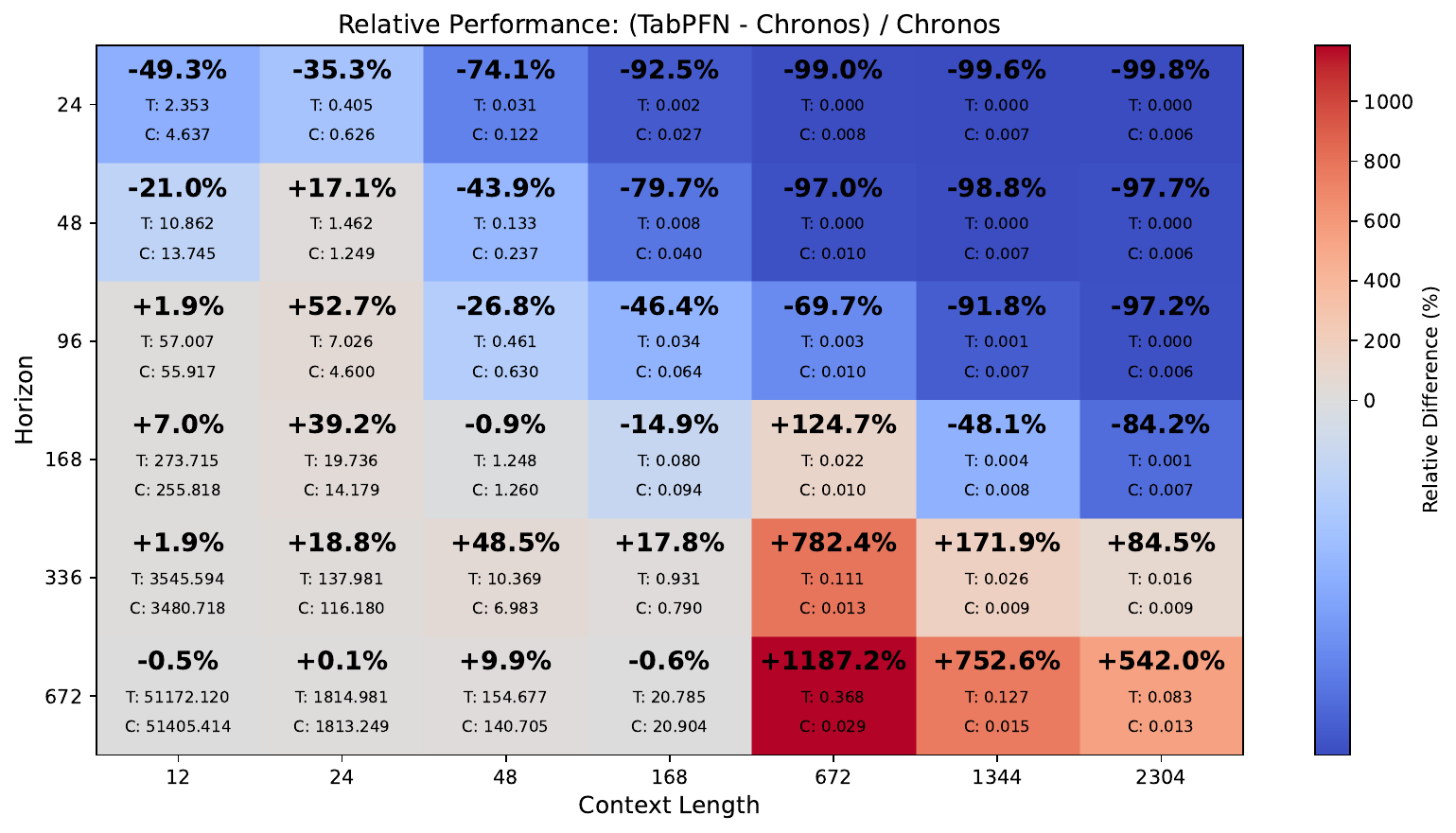}
\caption{Heatmaps of the relative performance of \texttt{Chronos-2} (C) compared to \texttt{TabPFN-TS} (T) for the sum experiment, expressed as $100*(T-C)/C$). \textcolor{blue}{Blue} indicates \textcolor{blue}{\texttt{TabPFN-TS}} outperforms \textcolor{red}{\texttt{Chronos-2}}, and \textcolor{red}{red} indicates \textcolor{red}{\texttt{Chronos-2}} outperforms \textcolor{blue}{\texttt{TabPFN-TS}}. Left: Random Walk dataset, Right: KernelSynth dataset.}
    \label{fig:sum_synth}
\end{figure}

\paragraph{Results for the sum experiment.} \label{par:sum_exp}Figure~\ref{fig:sum_synth} reports the relative performance of each model for the sum experiment, across both synthetic datasets. \texttt{TabPFN-TS} outperforms \texttt{Chronos-2} on short forecasting horizons and achieves comparable performance in several other settings. We hypothesize that \texttt{TabPFN-TS} performs better on short horizons because input and output features will be more similarly distributed, resulting in a task that more closely resembles the i.i.d problems encountered during pretraining.

\clearpage
\paragraph{Results for the aggregate experiment.} Table~\ref{tab:aggregate}, reports the errors of each model before and after providing the covariates. Similarly to the identity experiment, we observe that \texttt{Chronos-2} is a stronger univariate forecaster, but \texttt{TabPFN-TS} better captures the simple covariate-target relationship. Interestingly, removing the time embeddings and using only covariate features (\texttt{TabPFN}(cov) model) does not significantly degrade the performance, and becomes the most efficient model in the short-range setting.

\begin{table}[h!]
\caption{Normalized MSEs for \texttt{Chronos-2} and \texttt{TabPFN-TS} before and after providing covariates (cov). \texttt{TabPFN}(+cov) corresponds to \texttt{TabPFN-TS}(+cov) without the time embeddings.}
\vspace{1mm}
\centering
\scalebox{0.6}{
\begin{tabular}{cccccc}
\toprule
 Setting & \texttt{Chronos-2} &  \texttt{Chronos-2}(+cov) &  \texttt{TabPFN-TS} & \texttt{TabPFN-TS}(+cov) & \texttt{TabPFN}(cov) \\
\midrule
1344-336 & 0.1880 & 0.1940 & 0.2375 & \textbf{0.0008} & 0.0013 \\
672-168 & 0.1860 & 0.1905 & 0.2106 & \textbf{0.0013} & 0.0020 \\
168-24 & 0.1861 & 0.1999 & 0.3608 & \textbf{0.0010} & 0.0011 \\
24-24 & 0.5580 & 0.5625 & 1.6802 & 0.1160 & \textbf{0.0950} \\
\bottomrule
\end{tabular}
}
\label{tab:aggregate}
\end{table}

\begin{figure}[!h]
    \centering
    \includegraphics[width=0.45\textwidth]{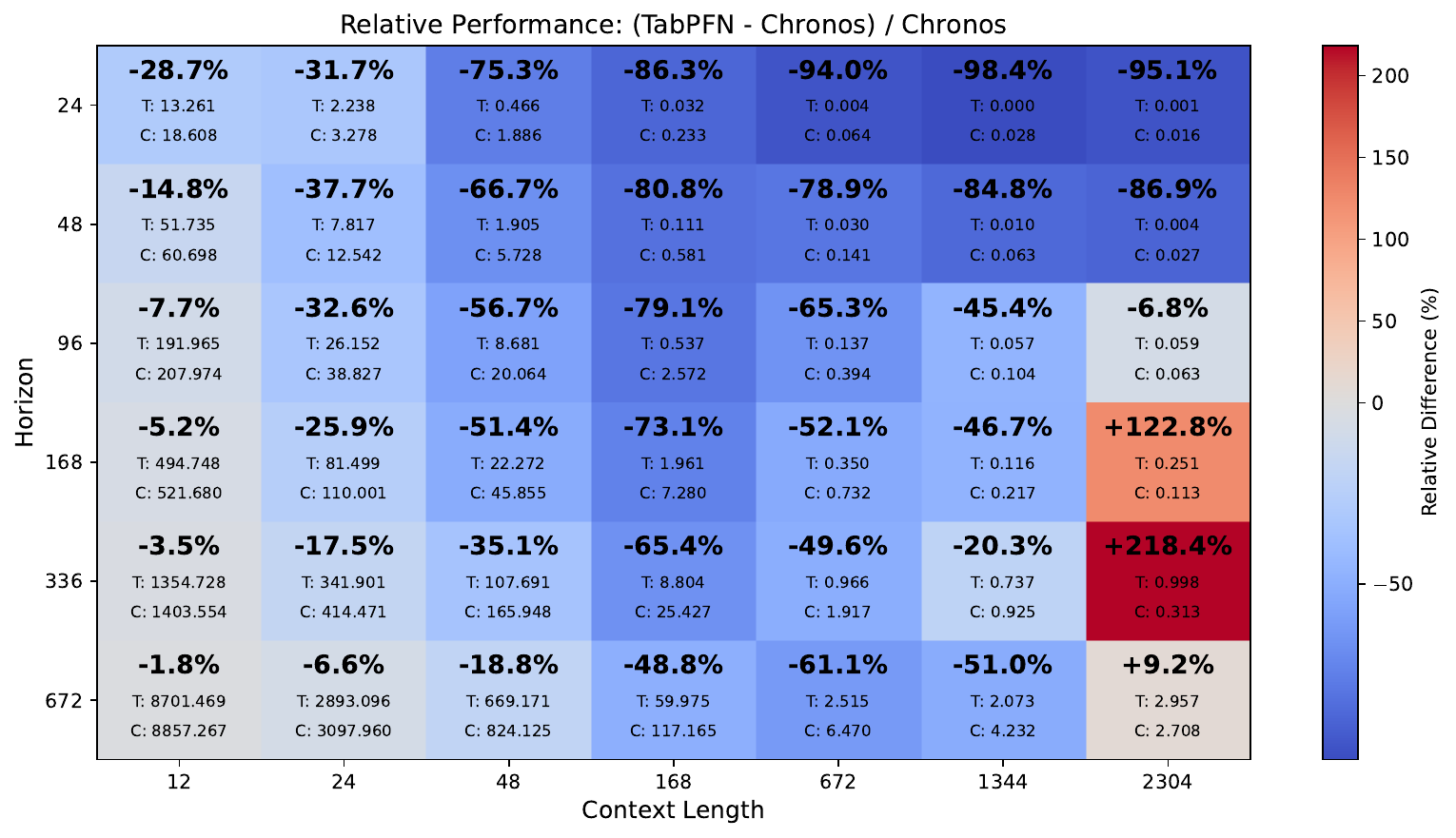}
    \includegraphics[width=0.45\textwidth]{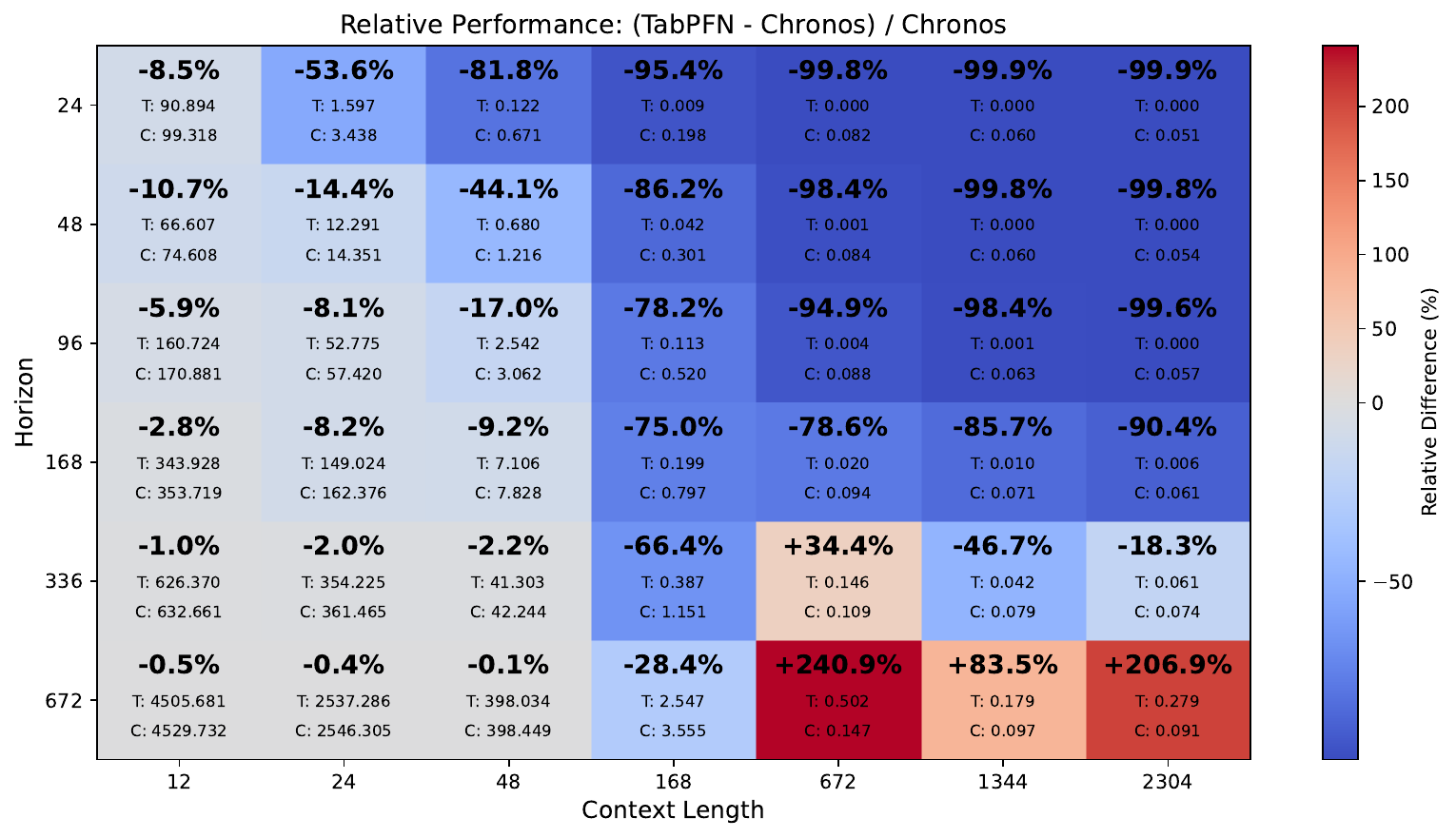}
\caption{Heatmaps of the relative performance of \texttt{Chronos-2} (C) compared to \texttt{TabPFN-TS} (T) for the quadratic experiment, expressed as $100*(T-C)/C$). \textcolor{blue}{Blue} indicates \textcolor{blue}{\texttt{TabPFN-TS}} outperforms \textcolor{red}{\texttt{Chronos-2}}, and \textcolor{red}{red} indicates \textcolor{red}{\texttt{Chronos-2}} outperforms \textcolor{blue}{\texttt{TabPFN-TS}}. Left: Random Walk dataset, Right: KernelSynth dataset.}
    \label{fig:pol_synth}
\end{figure}

\paragraph{Results for the quadratic experiment.} \label{par:quadratic_exp}Figure~\ref{fig:pol_synth} reports the results for the quadratric experiment. Once again, \texttt{TabPFN-TS} outperforms \texttt{Chronos-2} on the majority of settings, and particularly for short forecasting horizons.

\paragraph{Auto-regressive experiment} Additionally to the previous experiments, we ran an experiment with a target $Z=a+bX_t+cY_t^2+\phi Z_{t-1}$. Results are included in Appendix \ref{apx:autoregressive}.
By progressively increasing $\phi$, we increase the importance of time-dependence over the target-covariates relationship, which enables \texttt{Chronos-2} to better perform against \texttt{TabPFN-TS}. This suggests that the relative performance of both models depends on the relative importance of the time-to-target and covariate-to-target relationships for a given task.

\section{Conclusions}

In this paper, we investigated how \texttt{Chronos-2} and \texttt{TabPFN-TS} leverage covariate information to improve forecasting performance, by evaluating their performance on simple target-covariates tasks. Although \texttt{Chronos-2} achieves stronger performance on large-scale benchmarks such as \texttt{fev-bench}, our experiments show that \texttt{TabPFN-TS} demonstrates superior results on our simple synthetic experiments, especially for short forecasting horizons. We hypothesize that this advantage stems from its pretraining based on expressive prior and extensive synthetic regression problems.
This conclusion highlights a promising direction for future research in the development of TSFMs. In particular, a foundation architecture that combines explicit temporal dependency modeling, as in \texttt{Chronos-2}, with in-context regression capabilities inspired by \texttt{TabPFN-TS}, could represent an interesting direction for covariate-aware time series foundation models.

\clearpage

\section*{Acknowledgments}
Supported by EDF R\&D. Mariia Baranova, Adrien Petralia and Themis Palpanas are also supported by EU Horizon project DataGEMS ($101188416$), and by $Y \Pi AI
\Theta A$ \& NextGenerationEU project HARSH ($Y\Pi 3TA-0560901$) that is
carried out within the framework of the National Recovery and Resilience
Plan “Greece 2.0” with funding from the European Union –
NextGenerationEU.

\bibliography{iclr2026_conference}
\bibliographystyle{iclr2026_conference}

\appendix

\clearpage

\section{Experimental protocol}
\label{apx:protocol}

\paragraph{Datasets.} We consider three real-world datasets covering domains of energy, transport and climate: \textsc{Electricity} \citep{uci_electricity}, \textsc{Traffic} \citep{pems_data}, \citep{lai2018lstnet} and \textsc{Solar} \citep{nrel_solar}. \textsc{Solar} and \textsc{Electricity} are aggregated hourly (sum) and a dozen of users with many missing values are removed from \textsc{Electricity}. For the sum and quadratic experiments, we generate two synthetic datasets: one with the KernelSynth method, which was used to pretrain \texttt{Chronos-2} \citep{ansari2024chronos2}, and the latter with a random walk process. For the aggregate experiment, we manually build a \textsc{Conso} dataset: $\text{Conso}=\sqrt{\text{Traffic}} +\text{Electricity} - \text{Solar}^2$ (one time series for each user). For the quadratic experiment, the target is $Z=5 + 10X + 20Y^2$.

\paragraph{Setting.} We consider the usual look-back window to horizon forecasting setting, with varying look-back $L$ and horizon $H$ window sizes:
$$f: x\in \mathbb{R}^L \mapsto \hat{y}\in \mathbb{R}^H$$
Optionally, covariates $c \in \mathbb{R}^{L+H}$ can be included as inputs.

Our metric for evaluation is the normalized MSE, which corresponds to the MSE computed on predictions and ground-truths, normalized \textit{by instance}. That is:
$$L_f(x,y) = ||\hat{y}-\tilde{y}||_2^2\,, \quad \textrm{with}\quad \hat{y}=f(\tilde{x}),\; \tilde{x}=\frac{x-\mu_x}{\sigma_x},\;\tilde{y}=\frac{y-\mu_x}{\sigma_x}.$$

During evaluation, and for all $L-H$ settings, we apply a stride of 512 to obtain non overlapping and non periodic windows. We evaluate deterministically over those windows and for all individuals. 

\paragraph{Embeddings of \texttt{TabPFN}.}
Inspired by \texttt{TabPFN-TS} \citep{hoo2024tables2time}, we encode time indices for \texttt{TabPFN} with periodic time embeddings:
$$\phi(t\in [0,L+H])=\left(\frac{t}{L}, \cos\left(2\pi \frac{t}{T_i}\right),\sin\left(2\pi \frac{t}{T_i}\right), \dots\right) \quad T_i \in \text{\{Chosen periods\}}$$
We use daily ($T=24$) and weekly ($T=168$) periods for our real-world datasets, since all are hourly and are expected to follow these periods. This corresponds to $5$ time index features. When covariates are available as context, we encode them as features as well. When this is the case, we may remove the time embeddings. In our experiments with synthetic covariates, we observed better results without time embeddings, using only covariates used as features. Thus, \texttt{TabPFN-TS} from the synthetic experiments (results on Figure \ref{fig:sum_synth} and Figure \ref{fig:pol_synth}) does not include time embeddings, and is consequently identical to the original \texttt{TabPFN}. 

\section{Experiment with auto-regressive term}
\label{apx:autoregressive}
In complement to the quadratic experiment (Paragraph  \ref{par:quadratic_exp}), the following experiment investigates the effect of introducing temporal dependence into a nonlinear regression task. 
\paragraph{Settings.}The target variable is generated according to $ Z_t = a + b X_t + c Y_t^2 + \phi Z_{t-1}$, where \(X_t\) and \(Y_t\) are synthetically generated time series and \(a, b, c\) are fixed coefficients, as in the quadratic experiment (See Appendix ~\ref{apx:protocol}). The auto-regressive coefficient \(\phi\) is taken in the range \(\{0, 0.2, 0.4, 0.6, 0.8\}\), allowing a smooth transition from a purely instantaneous nonlinear relationship (\(\phi = 0\)) to a temporally dependent process.

\begin{figure}[H]
    \centering

    \begin{subfigure}{0.32\textwidth}
        \includegraphics[width=\linewidth]{images/heatmap_kernsynth_notimeembed.pdf}
        \caption{$\phi = 0$}
    \end{subfigure}
    \hfill
    \begin{subfigure}{0.32\textwidth}
        \includegraphics[width=\linewidth]{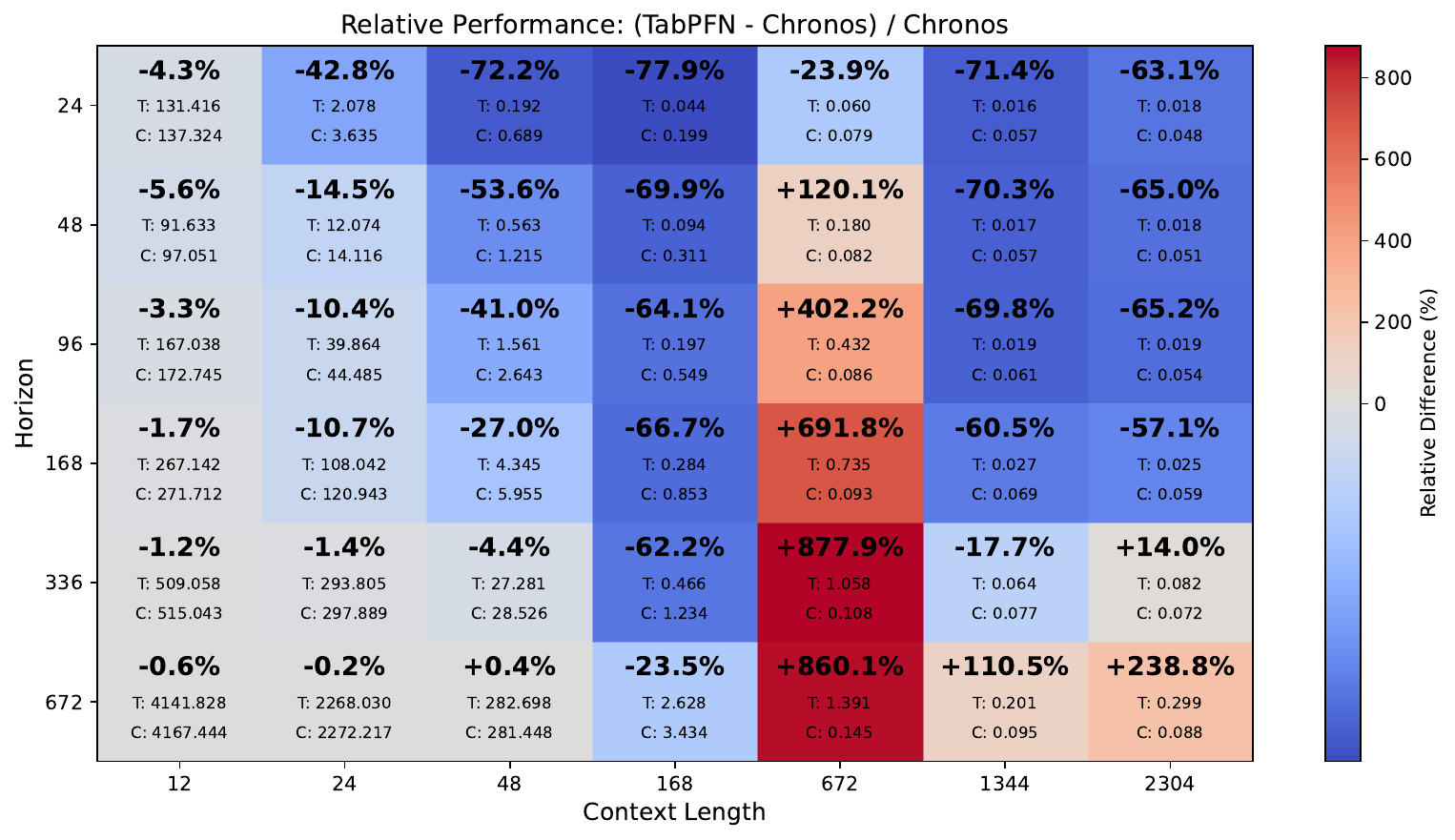}
        \caption{$\phi = 0.2$}
    \end{subfigure}
    \hfill
    \begin{subfigure}{0.32\textwidth}
        \includegraphics[width=\linewidth]{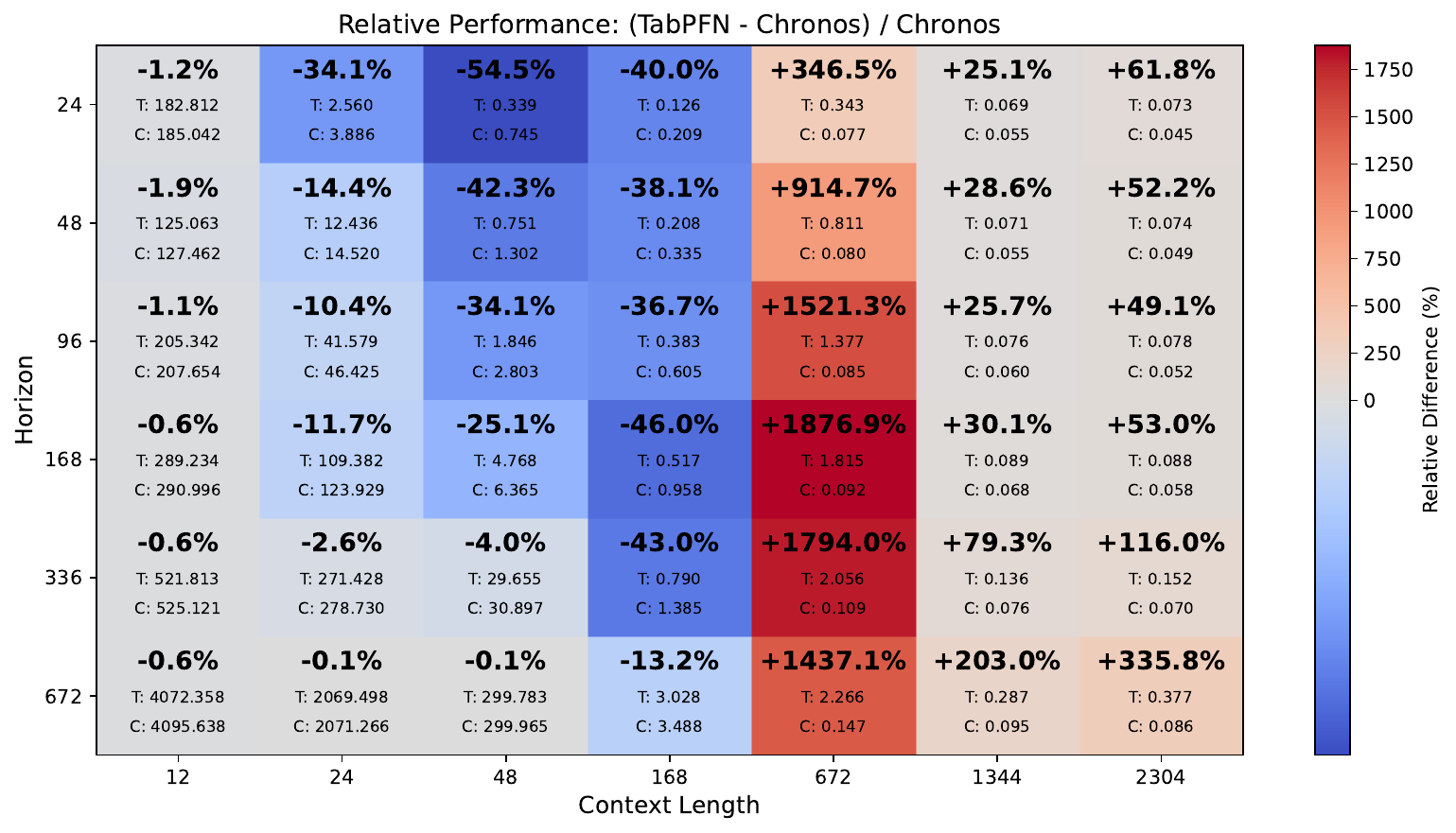}
        \caption{$\phi = 0.4$}
    \end{subfigure}

    \vspace{0.4cm}

    \begin{subfigure}{0.32\textwidth}
        \includegraphics[width=\linewidth]{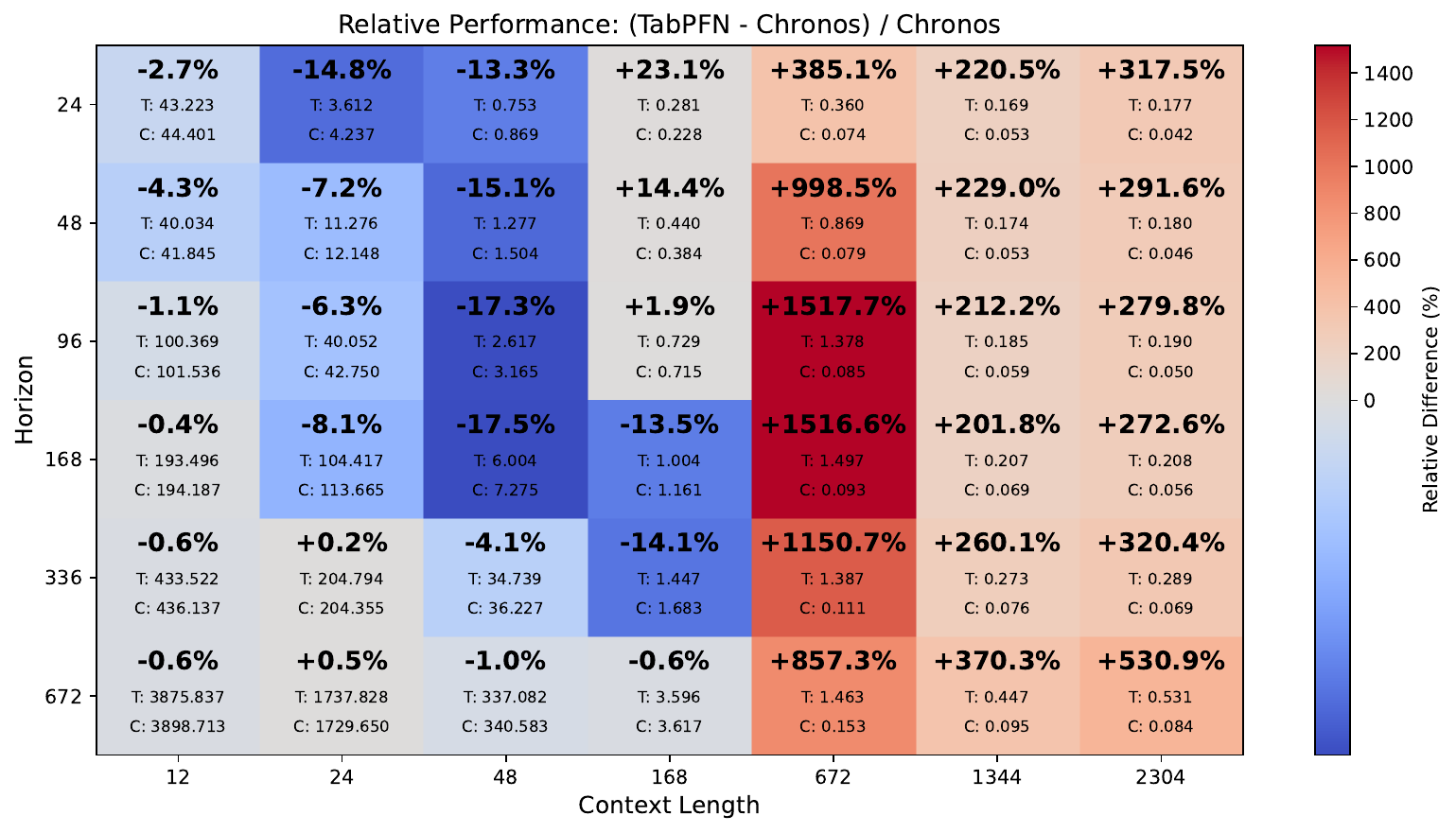}
        \caption{$\phi = 0.6$}
    \end{subfigure}
    \hfill
    \begin{subfigure}{0.32\textwidth}
        \includegraphics[width=\linewidth]{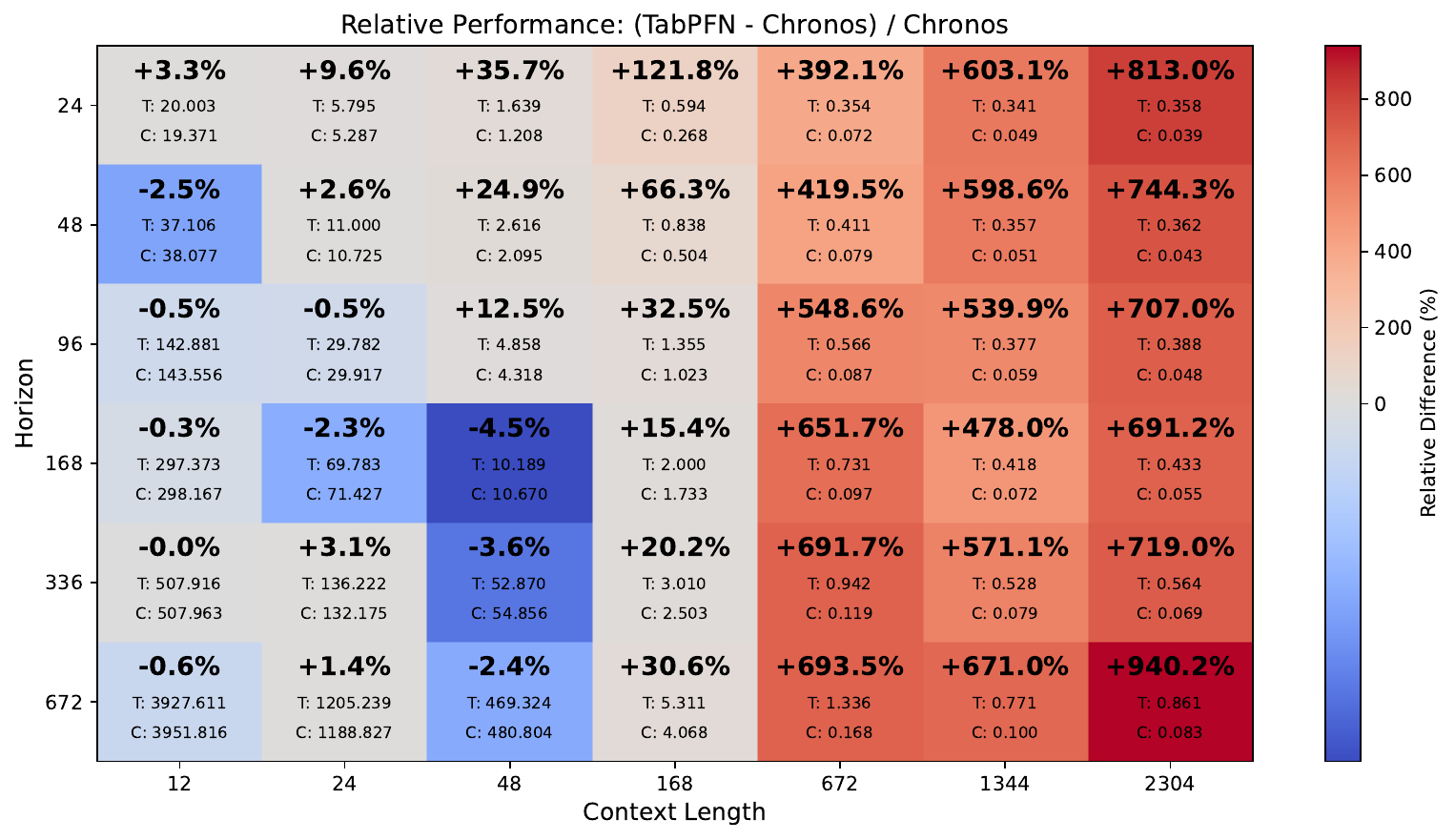}
        \caption{$\phi = 0.8$}
    \end{subfigure}
        \begin{subfigure}{0.32\textwidth}
        \includegraphics[width=\linewidth]{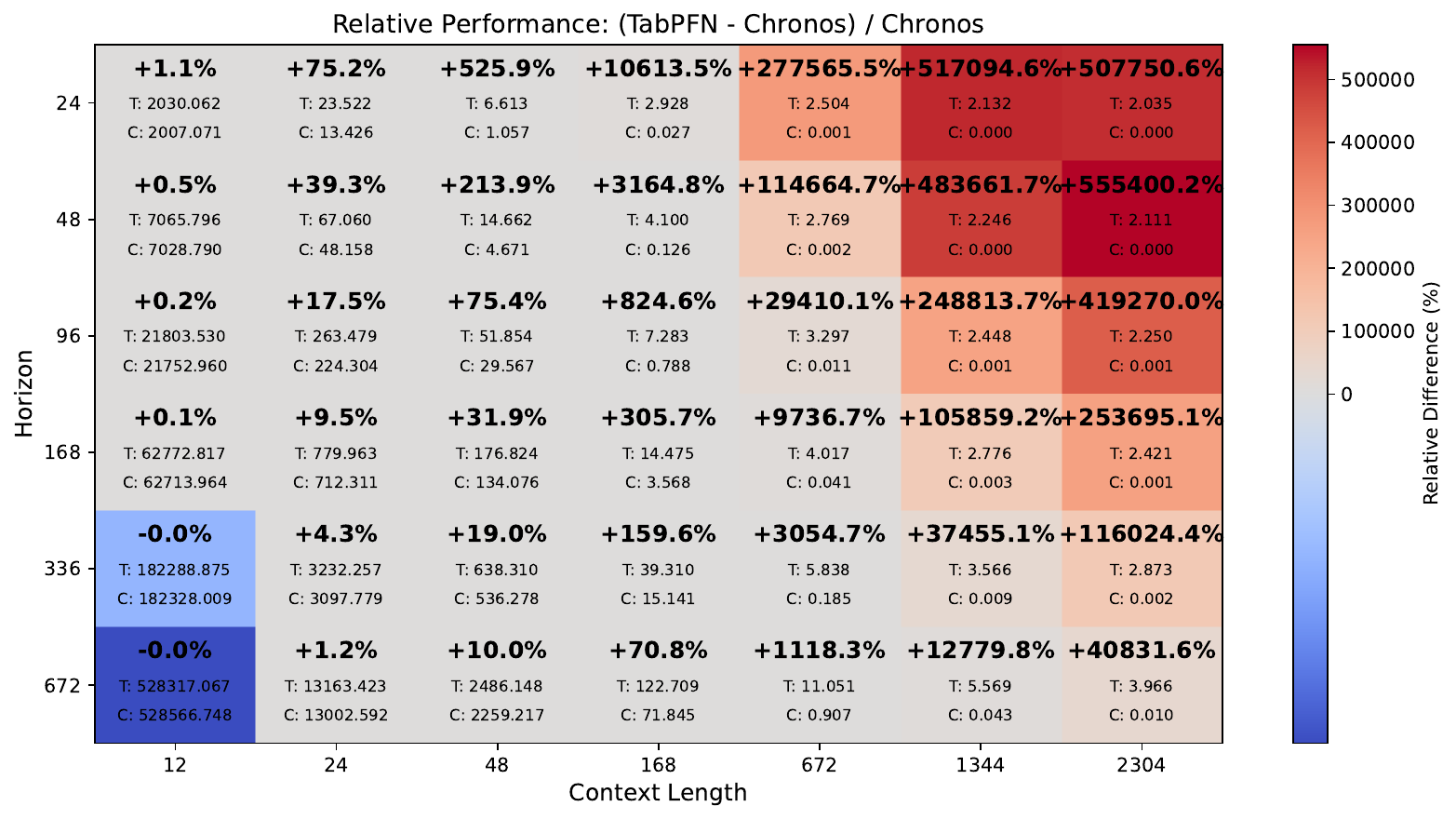}
        \caption{$\phi = 1$}
    \end{subfigure}

    \caption{
    Heatmaps showing the relative performance of \texttt{Chronos-2} (C) compared to \texttt{TabPFN-TS} (T), expressed as $100*(T-C)/C$), across context lengths and horizons for different values of $\phi$ on the KernelSynth dataset.
    \textcolor{blue}{Blue} regions indicate cases where \textcolor{blue}{\texttt{TabPFN-TS}} outperforms \textcolor{red}{\texttt{Chronos-2}}, while \textcolor{red}{red} regions indicate the opposite.
    }
    \label{fig:heatmaps_phi_kernelsynth_notime}
\end{figure}

\begin{figure}[H]
    \centering

    \begin{subfigure}{0.32\textwidth}
        \includegraphics[width=\linewidth]{images/heatmap_randwalk_notimeembed.pdf}
        \caption{$\phi = 0$}
    \end{subfigure}
    \hfill
    \begin{subfigure}{0.32\textwidth}
        \includegraphics[width=\linewidth]{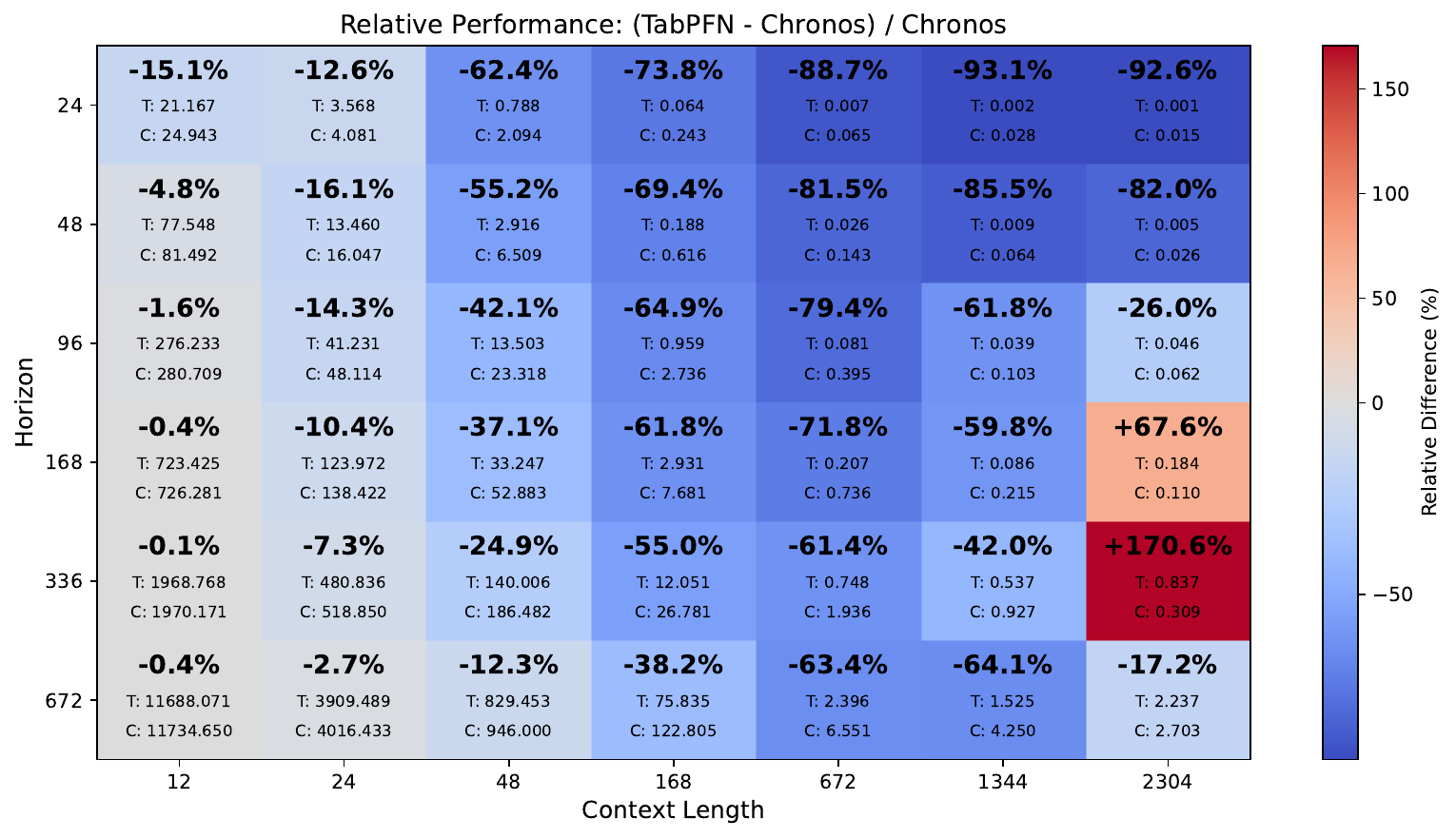}
        \caption{$\phi = 0.2$}
    \end{subfigure}
    \hfill
    \begin{subfigure}{0.32\textwidth}
        \includegraphics[width=\linewidth]{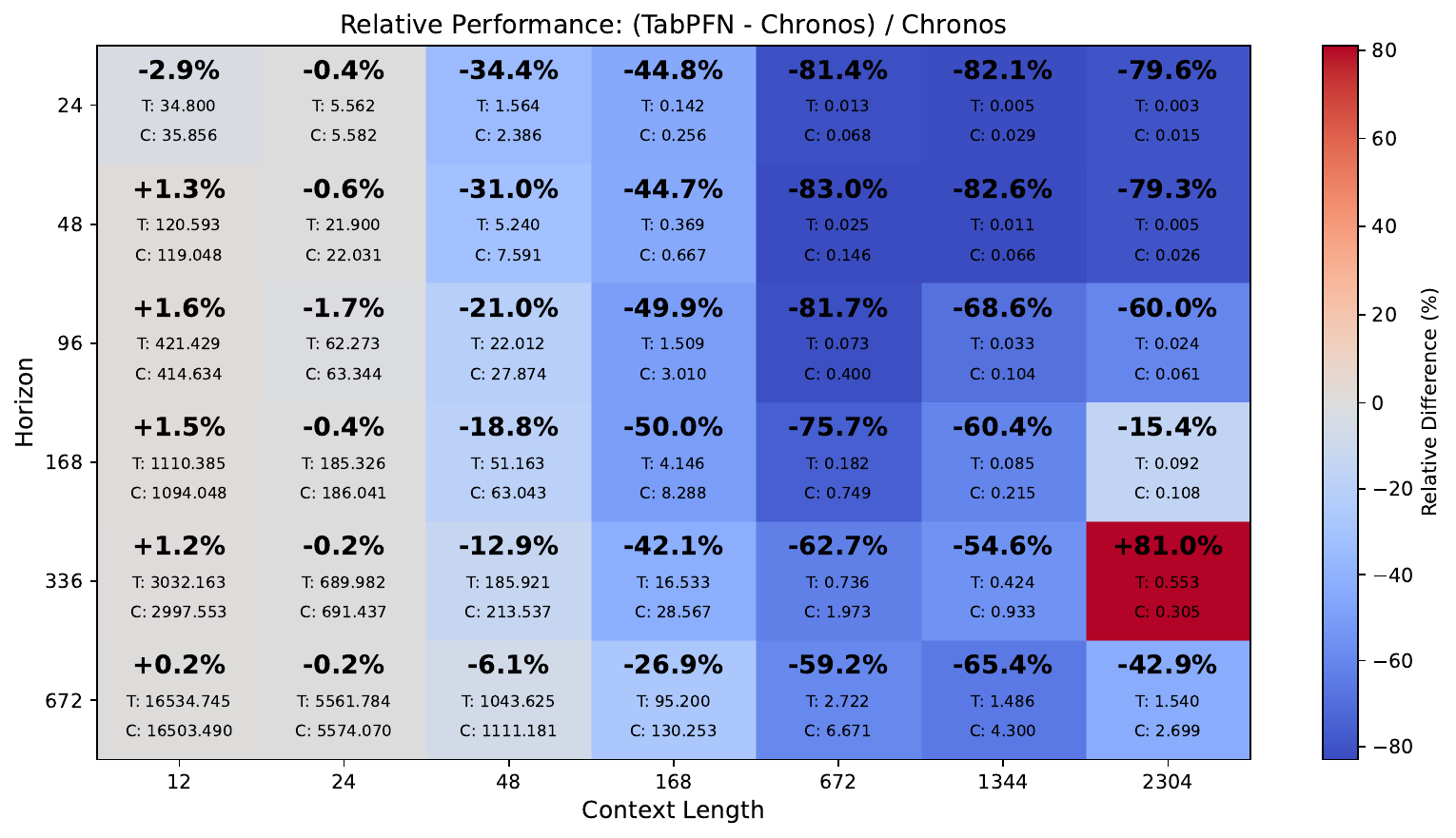}
        \caption{$\phi = 0.4$}
    \end{subfigure}

    \vspace{0.4cm}

    \begin{subfigure}{0.32\textwidth}
        \includegraphics[width=\linewidth]{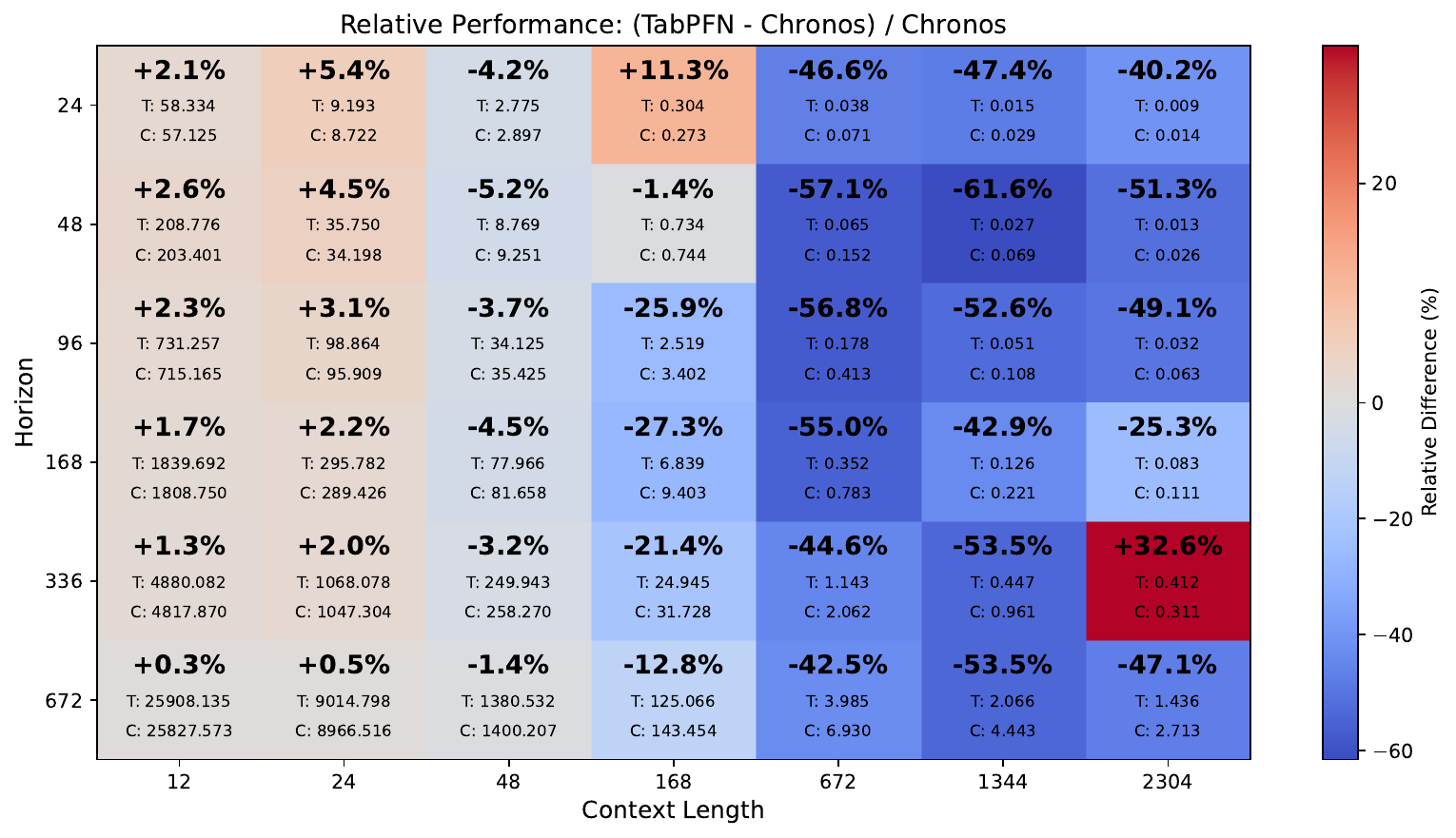}
        \caption{$\phi = 0.6$}
    \end{subfigure}
    \hfill
    \begin{subfigure}{0.32\textwidth}
        \includegraphics[width=\linewidth]{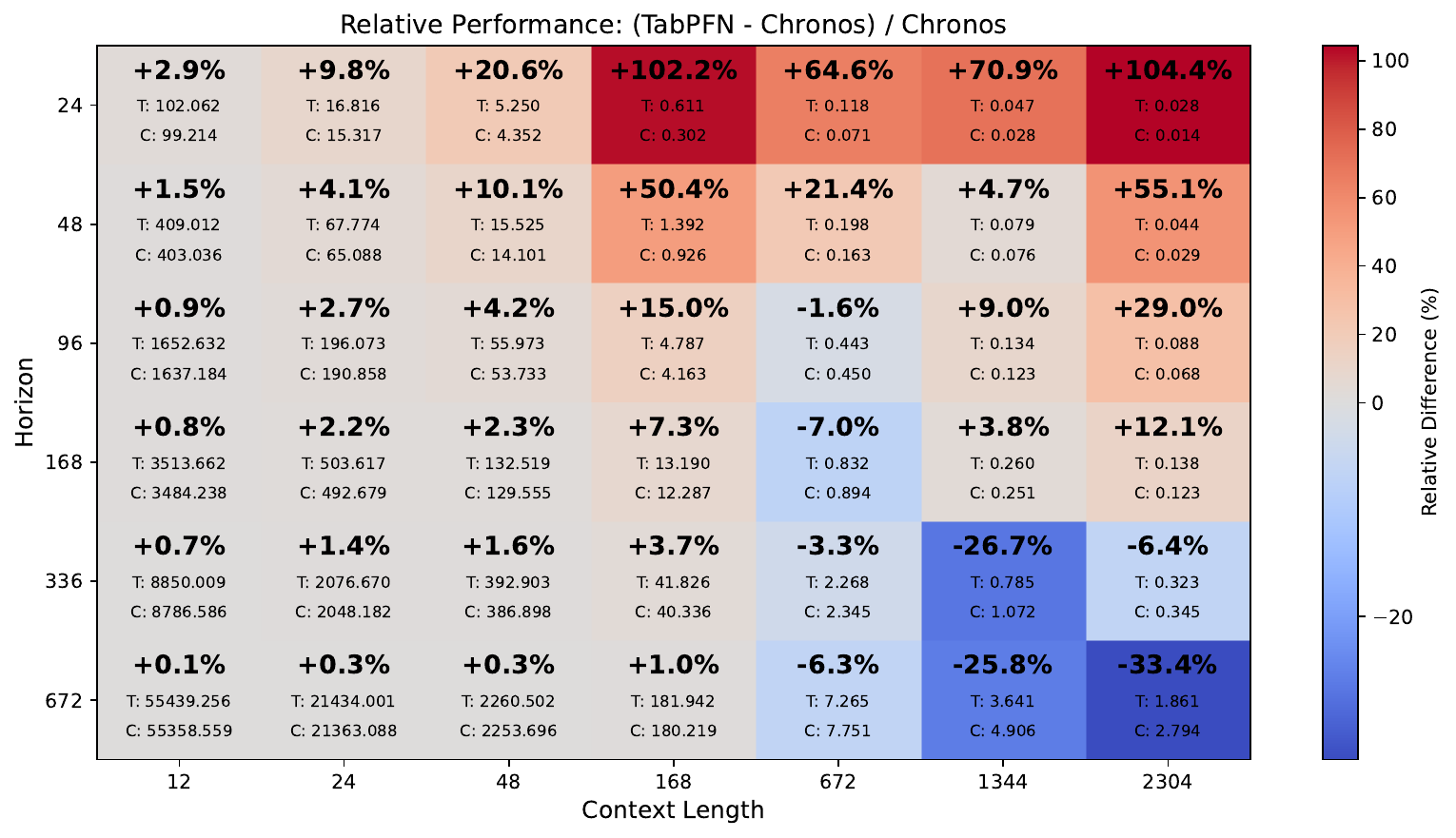}
        \caption{$\phi = 0.8$}
    \end{subfigure}
   \hfill
        \begin{subfigure}{0.32\textwidth}
        \includegraphics[width=\linewidth]{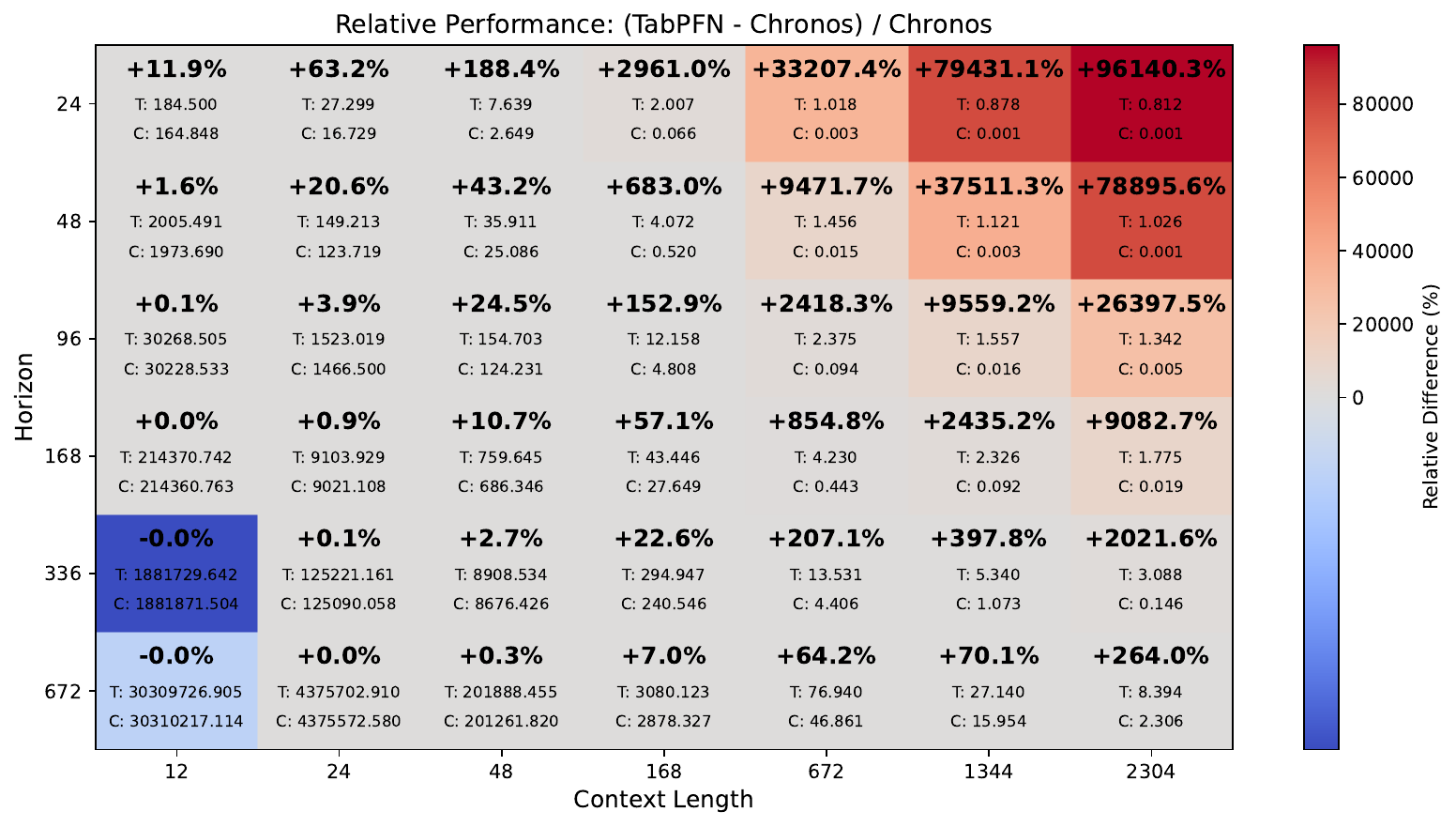}
        \caption{$\phi = 1$}
    \end{subfigure}

    \caption{
    Heatmaps showing the relative performance of \texttt{Chronos-2} (C) compared to \texttt{TabPFN-TS} (T), expressed as $100*(T-C)/C$), across context lengths and horizons for different values of $\phi$ on the Random Walk dataset.
    \textcolor{blue}{Blue} regions indicate cases where \textcolor{blue}{\texttt{TabPFN-TS}} outperforms \textcolor{red}{\texttt{Chronos-2}}, while \textcolor{red}{red} regions indicate the opposite.
    }
    \label{fig:heatmaps_phi_randwalk_notime}
\end{figure}

\paragraph{Results.} Figures~\ref{fig:heatmaps_phi_kernelsynth_notime} and \ref{fig:heatmaps_phi_randwalk_notime} showcase that as $\phi$ increases, the task transitions from a purely nonlinear tabular regression setting, where \texttt{TabPFN-TS} performs best
, to a temporally dependent forecasting problem in which \texttt{Chronos-2} increasingly outperforms \texttt{TabPFN-TS}  by exploiting the growing temporal structure. This suggests that the relative performance of both models depends on the relative importance of the time-to-target and covariate-to-target relationships for a given task.

\end{document}